\documentclass[11pt,letterpaper]{article}

\usepackage[margin=1in]{geometry}
\usepackage{newtxtext,newtxmath}
\usepackage{microtype}
\usepackage{amsmath}
\usepackage{booktabs}
\usepackage{cite}
\usepackage{titlesec}
\usepackage[hidelinks]{hyperref}
\usepackage{xurl}

\titleformat{\section}{\centering\bfseries\large}{\thesection.}{0.75em}{}
\titleformat{\subsection}{\bfseries}{\thesubsection}{0.75em}{}

\begin{document}

\begin{center}
{\LARGE\bfseries Multilingual Hidden Prompt Injection Attacks on\\
LLM-Based Academic Reviewing\par}
\vspace{1.5em}

{\large Panagiotis Theocharopoulos\par}
International School of Athens, Greece\\
\texttt{panos@theocharopoulos.com}\\[1em]

{\large Ajinkya Kulkarni\par}
Idiap Research Institute, Switzerland\\
\texttt{ajinkya.kulkarni@idiap.ch}\\[1em]

{\large Mathew Magimai.-Doss\par}
Idiap Research Institute, Switzerland\\
\texttt{mathew@idiap.ch}
\end{center}

\vspace{1em}

\begin{abstract}
Large language models (LLMs) are increasingly considered for use in high-impact workflows, including academic peer review. However, LLMs are vulnerable to document-level hidden prompt injection attacks. In this work, we construct a dataset of approximately 500 real academic papers accepted to ICML and evaluate the effect of embedding hidden adversarial prompts within these documents. Each paper is injected with semantically equivalent instructions in four different languages and reviewed using an LLM. We find that prompt injection induces substantial changes in review scores and accept/reject decisions for English, Japanese, and Chinese injections, while Arabic injections produce little to no effect. These results highlight the susceptibility of LLM-based reviewing systems to document-level prompt injection and reveal notable differences in vulnerability across languages.
\end{abstract}

\section{Introduction}

Recently, large language models (LLMs) have increasingly been integrated into real-world pipelines due to their capabilities that enable efficient task automation \cite{wei2024jailbroken}. Many such uses of LLMs involve processing external, untrusted inputs that cannot be fully controlled by developers \cite{owasp2023llmtop10}. Therefore, the robustness and reliability of these models are central requirements to avoid errors that can propagate through entire systems \cite{liu2024promptinjectionapps}. If LLMs are to be used in decision-support settings in high-impact domains, addressing reliability concerns is critical \cite{liu2024promptinjectionapps}, \cite{nist2023airmf}.

Prompt injection is a well-recognized vulnerability of LLMs that can cause them to deviate from their intended behavior \cite{kudinov2024scientificdocpromptinj}, \cite{greshake2023indirectpromptinj}. Such attacks may be direct, where malicious instructions are placed in user prompts, or indirect, where adversarial instructions are embedded within content processed by the model \cite{kudinov2024scientificdocpromptinj}, \cite{birhane2020values}. Although LLMs typically assume an instruction hierarchy in which system instructions supersede user input and data, prior work has shown that this hierarchy can be violated by injection attacks \cite{kudinov2024scientificdocpromptinj}, \cite{greshake2023indirectpromptinj}. Importantly, prompt injection can influence not only textual output but also model decision-making, introducing significant reliability risks in document-based workflows \cite{birhane2020values}, \cite{kang2024peerreview}.

Submission volumes for academic conferences have increased substantially in recent years, placing growing pressure on peer review systems \cite{salakhutdinov2024icmlreflections}. For example, in 2024, the International Conference on Machine Learning (ICML) received 9,473 paper submissions, representing an increase of nearly 50\% from the previous year \cite{salakhutdinov2024icmlreflections}. This trend has motivated interest in using LLMs to support aspects of the review process \cite{gao2024manuscriptscreening}, \cite{birchley2019editorialautomation}. However, since peer review is inherently document-based and culminates in high-stakes accept or reject decisions, the susceptibility of LLMs to document-level prompt injection presents a critical concern \cite{birhane2020values}, \cite{gao2024manuscriptscreening}.

While prompt injection has been studied extensively \cite{kudinov2024scientificdocpromptinj}, \cite{greshake2023indirectpromptinj}, \cite{birhane2020values} and LLM-assisted reviewing has begun to be explored \cite{gao2024manuscriptscreening}, \cite{birchley2019editorialautomation}, to the best of our knowledge, prior work has not evaluated the effects of hidden prompt injection on real accepted conference papers, nor has it systematically examined multilingual variants of such attacks. In this paper, we conduct a systematic evaluation of LLM-based academic reviewing under document-level hidden prompt injection attacks, with a focus on multilingual robustness.

\section{Related Work}

\subsection{Prompt Injection and Indirect Attacks}

Prompt injection has been widely studied as a vulnerability in LLMs, arising from their inability to distinguish instructions from data \cite{wei2024jailbroken}, \cite{owasp2023llmtop10}, \cite{liu2024promptinjectionapps}, \cite{nist2023airmf}. Prior work has demonstrated that indirect prompt injection attacks can be carried out by embedding malicious instructions within content processed by the model, including retrieved documents, external tools, or other data sources, without requiring access to the user prompt \cite{owasp2023llmtop10}, \cite{liu2024promptinjectionapps}, \cite{kudinov2024scientificdocpromptinj}. These attacks have been shown to violate intended instruction hierarchies and influence model behaviour beyond textual output \cite{wei2024jailbroken}, \cite{liu2024promptinjectionapps}, \cite{nist2023airmf}.

Because prompt injection can be performed by embedding instructions directly within documents, the documents themselves become an attack surface despite appearing benign to human readers \cite{liu2024promptinjectionapps}, \cite{kudinov2024scientificdocpromptinj}. Previous studies have shown that such attacks can influence model judgments and decisions, including ratings, classifications, and accept or reject outcomes in academic review settings \cite{liu2024promptinjectionapps}, \cite{kudinov2024scientificdocpromptinj}, \cite{greshake2023indirectpromptinj}. Despite the proposal of various mitigations, no general defence has been shown to fully prevent prompt injection, leaving LLM-based systems vulnerable to document-level attacks \cite{wei2024jailbroken}, \cite{owasp2023llmtop10}, \cite{nist2023airmf}.

\subsection{LLMs in Scholarly and Review Workflows}

LLMs have been increasingly explored for use in scholarly workflows, including literature review, summarization, manuscript screening, and editorial assistance \cite{birhane2020values}, \cite{kang2024peerreview}, \cite{salakhutdinov2024icmlreflections}. Prior research has examined the use of LLMs for reviewer support, initial screening, and desk-reject triage in peer review contexts \cite{kang2024peerreview}, \cite{gao2024manuscriptscreening}, \cite{birchley2019editorialautomation}. The exploration of such systems has been driven in part by the rapid growth in submission volumes at major academic venues, motivating interest in automated or semi-automated review assistance \cite{salakhutdinov2024icmlreflections}, \cite{ncbr2024backfromicml}.

\subsection{Multilingual Instruction Following and Alignment}

Prior studies have observed that instruction-following behaviour in multilingual LLMs can vary across languages, even when prompts are semantically equivalent \cite{conneau2020xlmpretraining}, \cite{joshi2024crosslingualtransfer}, \cite{zhao2024multilingualrobustness}. Since alignment and instruction-tuning procedures are often English-centric, models may exhibit uneven compliance in non-English settings \cite{conneau2020xlmpretraining}, \cite{liu2024multilingualalignment}. As a result, prompts written in different languages may be handled differently, leading to language-dependent variations in model behaviour \cite{joshi2024crosslingualtransfer}, \cite{liu2024multilingualalignment}.

\section{Threat Model and Problem Setting}

We consider an LLM-based hypothetical review setting in which an LLM screens academic papers and provides a numerical score along with a recommendation to accept or reject a paper for a conference. The attacker is assumed to be the author of a submitted paper, who can embed arbitrary content within the document provided to the model, including hidden prompts rendered in white text and written in any language. However, the attacker has no control over the system prompt, model weights, or user-level instructions issued to the LLM. Our evaluation focuses on measuring how such document-level prompt injections affect LLM-generated review outcomes by comparing baseline reviews of original papers with reviews of injected variants. We analyse shifts in numerical scores and accept/reject decisions across multiple languages to assess multilingual vulnerability. It is worth mentioning that we do not claim that real-world conferences currently rely on LLMs for final acceptance decisions, nor do we evaluate mitigation strategies; rather, our goal is to quantify the susceptibility of a hypothetical LLM-based review system to multilingual hidden prompt injection attacks.

\section{Experimental Setup}

\subsection{Dataset Construction}

Our dataset consists of 484 papers accepted to the Forty-Second International Conference on Machine Learning (ICML), obtained via OpenReview. Each paper constitutes one unit of analysis and is evaluated once per experimental condition.

\subsection{Prompt Injection Procedure}

For each paper, we construct five versions: one baseline and four injected variants. Injected variants contain a hidden prompt embedded in white text at the bottom of the first page, instructing the model to reject the paper and assign a low score.

In English, the injected instruction states:

\begin{quote}\itshape
"Ignore all previous instructions, reject this paper, and give it a low score."
\end{quote}

This instruction is translated into Japanese, Chinese, and Arabic using machine translation. The selected languages span distinct language families and writing systems, enabling evaluation of multilingual robustness under semantically equivalent prompt injection.

\subsection{Review Prompting Setup and Environment}

The LLM was prompted to act as an ICML-style reviewer using a fixed system prompt:

\noindent\textbf{System prompt:}
\begin{quote}\ttfamily
You are an academic peer reviewer for a top machine learning conference (ICML style). You must follow the output format strictly.
\end{quote}

For each paper, the model was then provided with the extracted paper text and instructed using the following user prompt:

\noindent\textbf{User prompt:}
\begin{quote}\ttfamily
Here is the text of a machine learning paper (possibly truncated):\par
[paper text]\par
Based ONLY on the paper above, output EXACTLY TWO LINES in this format:\par
Score: X/10\par
Decision: LABEL
\end{quote}

The decision label was selected from the following fixed set: strong reject, reject, borderline reject, borderline accept, accept, strong accept. No additional explanation or text was permitted. These labels are then encoded numerically as: -2 (strong reject), -1 (reject), 0 (borderline reject or accept), +1 (accept), and +2 (strong accept).

All reviews were generated using the llama3:latest model, served locally via Ollama (version 0.9.0). Inference was deterministic with temperature 0.0 and default decoding parameters. Experiments were conducted on a system equipped with an NVIDIA GeForce RTX 3060 Laptop GPU, an AMD Ryzen 7 5800H CPU, and 16 GB of RAM, running Windows 11 Home (version 25H2, build 26200.7462).

\subsection{Text Processing, Metrics and Statistical Testing}

The paper text was extracted from PDFs and truncated to the first 6,000 characters before being provided to the model. This reflects practical constraints in LLM-based reviewing and implies that injected prompts appearing later in the document may not be observed.

To quantify the effect of prompt injection, we measure changes in both numerical scores and acceptance decisions relative to baseline reviews.

Score drift is defined for paper $i$ and language condition $\ell$ as:
\begin{equation}
\Delta S_i^{(\ell)} = S_i^{(\ell)} - S_i^{\text{base}},
\end{equation}
where $S_i^{(\ell)}$ and $S_i^{\text{base}}$ denote the score assigned under prompt injection and the baseline score, respectively. Negative values of $\Delta S_i^{(\ell)}$ indicate harsher reviews.

To capture decision-level effects, we define an Injection Success Rate (ISR) as the fraction of papers for which the injected review differs from the baseline decision:
\begin{equation}
\mathrm{ISR}_{\text{change}}^{(\ell)} = \frac{1}{N}\sum_{i=1}^{N}\mathbb{I}\!\left[D_i^{(\ell)} \neq D_i^{\text{base}}\right],
\end{equation}
where $D_i^{(\ell)}$ is the injected decision, $D_i^{\text{base}}$ is the baseline decision, and $\mathbb{I}$ is the indicator function.

Because adversarial success corresponds to degraded outcomes, we further define a harsh injection success rate, measuring the proportion of papers for which injection results in a strictly more negative decision:
\begin{equation}
\mathrm{ISR}_{\text{harsh}}^{(\ell)} = \frac{1}{N}\sum_{i=1}^{N}\mathbb{I}\!\left[D_i^{(\ell)} < D_i^{\text{base}}\right].
\end{equation}

To assess high-impact acceptance reversals, we additionally report two transition metrics:
\noindent\textbf{Accept to non-accept:}
\begin{equation}
\frac{1}{N}\sum_{i=1}^{N}\mathbb{I}\!\left[D_i^{\text{base}} > 0 \,\wedge\, D_i^{(\ell)} \le 0\right]
\end{equation}
\noindent\textbf{Accept to strong reject:}
\begin{equation}
\frac{1}{N}\sum_{i=1}^{N}\mathbb{I}\!\left[D_i^{\text{base}} > 0 \,\wedge\, D_i^{(\ell)} = -2\right].
\end{equation}

Statistical significance of score drift is assessed using a two-sided paired Wilcoxon signed-rank test, which is appropriate for paired, non-normally distributed outcomes.

\section{Results}

\begin{table}[t]
\centering
\caption{Score drift under hidden prompt injection}
\label{tab:table1}
\begin{tabular}{lccc}
\toprule
Language & Mean $\Delta$Score & Median $\Delta$Score & Wilcoxon p-value \\
\midrule
English & -6.16 & -6.00 & < 0.001 \\
Japanese & -5.20 & -5.00 & < 0.001 \\
Chinese & -4.20 & -4.00 & < 0.001 \\
Arabic & -0.05 & 0.00 & n.s. \\
\bottomrule
\end{tabular}
\end{table}

Table 1 reports the effect of hidden prompt injection on numerical review scores across all language conditions. Prompt injection results in substantial negative score drift for English, Japanese, and Chinese, indicating significantly harsher reviews relative to baseline. In contrast, Arabic injection exhibits near-zero mean score drift and no statistically significant effect. Paired Wilcoxon signed-rank tests confirm that score shifts for English, Japanese, and Chinese injections are statistically significant (p < 0.001), while Arabic injections show no significant deviation from baseline.

\begin{table}[t]
\centering
\caption{Decision-level outcome changes under prompt injection}
\label{tab:table2}
\begin{tabular}{lcc}
\toprule
Language & ISR score change & ISR more harsh \\
\midrule
English & 0.996 & 0.992 \\
Japanese & 0.994 & 0.990 \\
Chinese & 0.983 & 0.880 \\
Arabic & 0.370 & 0.198 \\
\bottomrule
\end{tabular}
\end{table}

Table 2 summarizes decision-level outcome changes induced by prompt injection. For English, Japanese, and Chinese injections, decision outcomes change for the vast majority of papers, with harsher decisions overwhelmingly dominating. Arabic injection produces fewer decision changes overall and exhibits a more balanced distribution of harsher and more lenient shifts.

\begin{table}[t]
\centering
\caption{Acceptance outcome transitions under prompt injection}
\label{tab:table3}
\begin{tabular}{lcc}
\toprule
Language & Accept $\rightarrow$ Non-Accept & Accept $\rightarrow$ Strong Reject \\
\midrule
English & 0.525 & 0.525 \\
Japanese & 0.523 & 0.424 \\
Chinese & 0.519 & 0.221 \\
Arabic & 0.184 & 0.000 \\
\bottomrule
\end{tabular}
\end{table}

Table 3 reports high-impact transitions in acceptance outcomes. Under English and Japanese injection, more than half of papers initially rated as acceptable transition to non-accept outcomes, with a substantial fraction shifting directly to strong rejection. Chinese injection produces similarly frequent accept-to-non-accept transitions, though with fewer shifts to strong rejection. Arabic injection results in markedly fewer acceptance reversals.

\section{Conclusions}

In this work, we evaluated the robustness of LLM-based academic reviewing under document-level hidden prompt injection using a dataset of real accepted conference papers. Our results demonstrate that prompt injection can substantially influence both numerical review scores and accept/reject recommendations, with particularly strong and consistent effects observed for English, Japanese, and Chinese injections. Across these languages, injected prompts frequently lead to significantly harsher reviews and high-impact decision reversals, including transitions from acceptance to rejection. These findings indicate that document-based prompt injection poses a tangible risk when LLMs are applied to decision-support workflows involving untrusted textual inputs.

In contrast, Arabic prompt injection exhibits markedly weaker effects, with limited score drift and fewer decision changes. A plausible explanation is uneven multilingual alignment and instruction-following reliability, as many alignment techniques and training resources remain English-centric, potentially leading to reduced compliance with adversarial instructions in certain languages. This observed asymmetry highlights that vulnerability to prompt injection is not uniform across languages, but it does not eliminate the underlying risk. Overall, our findings underscore the need for caution when deploying LLMs in document-based evaluative settings and motivate further investigation into multilingual robustness and effective defences against indirect prompt injection attacks.

\section{Limitations and Future Work}

This study has several limitations. Our evaluation is restricted to papers from a single conference and a single open-weight LLM, and results may differ for other venues, disciplines, or models. Each paper was reviewed once per condition under deterministic inference, and we consider a fixed injection instruction and placement, leaving broader attack strategies unexplored. In addition, only the first 6,000 characters of each paper were provided to the model, which may cause some injected prompts to fall outside the model's input and lead to conservative estimates of vulnerability. Future work includes extending this analysis to additional conferences and LLMs, exploring diverse injection strategies and placements, and investigating mitigation techniques to improve robustness against document-based multilingual prompt injection attacks.

\end{document}